# SLAM

## Solutions lexicales automatiques pour métaphores

### Yann Desalle[1, 2], Bruno Gaume[1, 3], Karine Duvignau[1]

[1] CLLE-ERSS, [2] Octogone-Lordat
Université Toulouse 2 - 5, allée Antonio Machado
F - 31058 Toulouse - Cedex 9

[3] IRIT-RPDMP
Université Toulouse 3- 118, route de Narbonne
 F-31062 Toulouse - Cedex 9

{yann.desalle, gaume, duvignau}@univ-tlse2.fr

RÉSUMÉ.
*Cet article[1] présente SLAM, un modèle de solutions lexicales automatiques pour métaphores de type « déshabiller\*[2] une pomme ». SLAM recherche une solution conventionnelle à ce type de production métaphorique. Pour cela, il croise l'axe paradigmatique du verbe métaphorique « déshabiller\* », sur lequel est rapproché le verbe « peler », avec l'axe syntagmatique émergeant d'un corpus dans lequel « peler une pomme » est une structure syntaxiquement et sémantiquement régulière. Nous testons le modèle sur DicoSyn, un réseau de synonymes de type petit monde, pour le calcul de l'axe paradigmatique et sur le corpus Frantext.20 pour le calcul de l'axe syntagmatique. Enfin, nous évaluons ce modèle à l'aide d'un corpus expérimental issu de la base de données Flexsem.*

ABSTRACT.

*This article presents SLAM, an Automatic Solver for Lexical Metaphors like "déshabiller\* une pomme" (to undress\* an apple). SLAM calculates a conventional solution for these productions. To carry on it, SLAM has to intersect the paradigmatic axis of the metaphorical verb "déshabiller\*", where "peler" ("to peel") comes closer, with a syntagmatic axis that comes from a corpus where "peler une pomme" (to peel an apple) is semantically and syntactically regular. We test this model on DicoSyn, which is a "small world" network of synonyms, to compute the paradigmatic axis and on Frantext.20, a French corpus, to compute the syntagmatic axis. Further, we evaluate the model with a sample of an experimental corpus of the database of Flexsem*

MOTS-CLÉS : *métaphore, synonyme, proxsem, graphe, petit monde, prox, flexsem, syntex.*

---

## 1. Introduction

Cet article présente SLAM, un modèle de solutions lexicales automatiques pour métaphores du type « *déshabiller\* une pomme* ». Selon Aristote, (Aristote, *Poétique*), cette métaphore se bâtit à partir de l'analogie conceptuelle d'un quadruplet du type : PELER:POMME::DESHABILLER:POUPEE[3] : le quadruplet conceptuel $c_1$:$c_2$::$c_3$:$c_4$ est analogique[4] si le concept $c_1$ entretient avec le concept $c_2$ la même relation que le concept $c_3$ entretient avec le concept $c_4$ ($c_1$ est à $c_2$ ce que $c_3$ est à $c_4$) d'où PELER:POMME::DESHABILLER:POUPEE est analogique, car le concept PELER est au concept POMME ce que le concept DESHABILLER est au concept POUPEE.

Depuis les travaux de Lakoff et Johnson (Lakoff et Johnson, 1980), il est admis que le processus métaphorique prend sa source au niveau conceptuel. Ainsi, lorsque nous sommes face à un domaine conceptuel nécessitant une meilleure appréhension de sa structure, appelé domaine cible, nous projetons sur celui-ci la structure d'un domaine conceptuel source que nos expériences corporelles, sociales, culturelles... ont permis de consolider. Cette projection se fait à partir de l'identification de similitudes structurelles entre le domaine source et le domaine cible (Gentner, 1983 ; Gineste, 1997 ; Sander, 2003).

Un locuteur voulant communiquer un événement A – ex. : [l'action de PELER une POMME] – peut produire un énoncé conventionnel « *peler une pomme* » mais il peut aussi produire un énoncé métaphorique « *déshabiller une pomme* », ce qui est fréquent chez les jeunes enfants (Duvignau et Gaume, 2005) : l'enfant cherchant à communiquer un événement A – ex. : [l'action de PELER une POMME] – pour lequel il ne dispose pas encore de catégorie verbale constituée (1) fait une analogie avec un ancien événement B [l'action de DESHABILLER une POUPEE], déjà mémorisé avec l'entrée lexicale « *déshabiller* » et, (2) utilisant cette analogie, dit « *déshabiller une pomme* » pour communiquer l'événement A.

Afin d'exploiter dans notre modèle, à un niveau lexical, cette notion d'analogie conceptuelle, nous distinguons deux types de relations entre les lexèmes :

**(1) les relations syntagmatiques entre termes** : le terme $t_1$ est en relation syntagmatique avec le terme $t_2$ si le terme $t_1$ peut apparaître dans un nombre plus ou moins grand de contextes syntaxiques de $t_2$ ;

Exemple : le terme « *peler* » est en relation syntagmatique avec le terme « *pomme* ».

---

3. Dans cet article, nous noterons en petites capitales le niveau conceptuel (ex : POMME) et entre crochets les événements (ex : [l'action de PELER une POMME]), alors que nous noterons en italique et entre guillemets le niveau langagier (ex : « *peler une pomme* »).
4. Ce patron d'analogie conceptuelle est celui donné par Aristote lui-même, dans sa poétique.

**(2) les relations paradigmatiques entre termes :** le terme $t_1$ est en relation paradigmatique avec le terme $t_2$ si le terme $t_1$ peut se substituer au terme $t_2$, sans changement fondamental de sens, dans un nombre plus ou moins grand de contextes syntaxiques.

Exemple : le terme « *peler* » est en relation paradigmatique avec le terme « *éplucher* ».

Dans ce cadre, SLAM a pour but de résoudre lexicalement les métaphores analogiques (voir figure 1).

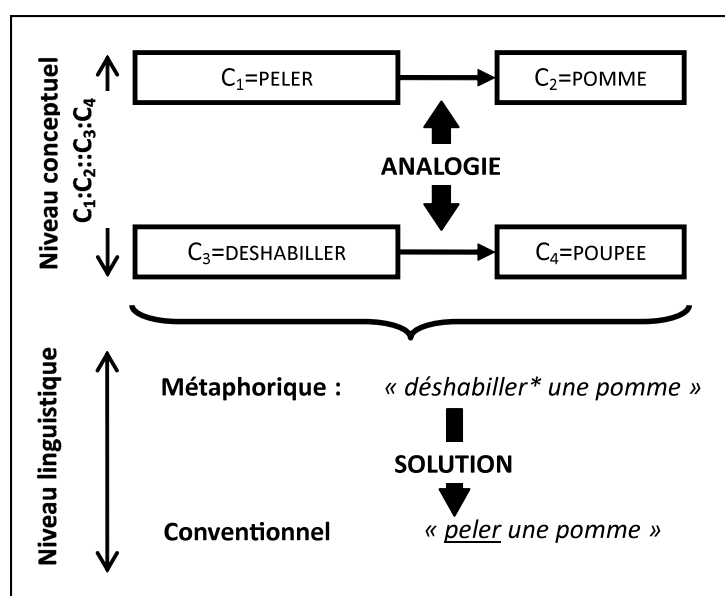

**Figure 1.** *Résoudre la métaphore analogique*

Dans la figure 1 ci-dessus, il existe au niveau conceptuel une relation d'analogie PELER:POMME::DESHABILLER:POUPEE, ce qui permet à un locuteur de résoudre au niveau linguistique la métaphore « *déshabiller\* une pomme* » par une solution conventionnelle[5] « *peler* » dans l'expression « *peler une pomme* ».

SLAM, qui prend en entrée une métaphore analogique quelconque comme « *déshabiller\* une pomme* », doit donc proposer en sortie une solution lexicale conventionnelle, comme « *peler* », de manière automatique, sans avoir accès[6] au niveau conceptuel. En effet, ce niveau serait très difficile à représenter formellement, de manière exhaustive, pour l'ensemble de tous les concepts et dans tous les domaines. Aussi, afin de compenser ce défaut de représentation du niveau conceptuel, SLAM repose sur le croisement de l'axe paradigmatique du foyer de la métaphore (ex. : « *déshabiller\** ») avec l'axe syntagmatique du terme créant la tension avec le foyer métaphorique (ex. : « *pomme* »). Concernant l'axe paradigmatique, SLAM s'appuie sur Prox afin de construire une similarité entre entités lexicales. Concernant l'axe syntagmatique, SLAM s'appuie sur l'analyseur syntaxique Syntex afin d'extraire les triplets syntaxiques <régisseur, relation, régi> présents dans un corpus avec leur effectif. Par ailleurs, il faut distinguer les métaphores créatives, qui s'appuient sur des analogies non systématisées, des métaphores conventionnelles dont l'utilisation plus fréquente a été mise au jour par (Lakoff et Johnson, 1980), ces dernières s'apparentant à de la polysémie. En effet, par un processus d'abstraction de son domaine sémantique d'origine, le foyer métaphorique – conventionnalisé par un usage fréquent de son sens figuré – appartient aux deux domaines sémantiques mis en jeu par une métaphore analogique (Gentner *et al.*, 2001). Ainsi, les métaphores conventionnelles ne génèrent que peu, et même parfois plus aucune tension contrairement aux métaphores créatives dont la vivacité entraîne une tension sémantique qu'il est intéressant de modéliser. C'est pourquoi nous appliquons SLAM aux métaphores créatives, la résolution d'une métaphore conventionnelle s'apparentant plutôt à la désambiguïsation d'un terme polysémique conventionnel.

---

5. Solution conventionnelle : qui n'entraîne aucune tension, ni d'un point de vue sémantique (le contexte lexicale de la solution dans l'expression) ni d'un point de vue pragmatique (voir note 7 sur la troponymie).

Dans la suite de cet article, nous présentons le fonctionnement de SLAM et les résultats d'une première évaluation. À la section 2, nous introduisons le cadre théorique détaillé de ces travaux en nous focalisant au § 2.2 sur les modèles computationnels existants autour de la métaphore. Avant de définir formellement SLAM au § 3.3, nous décrivons, d'une part, au § 3.1, Syntex, qui est un analyseur syntaxique à partir duquel SLAM calcule l'axe syntagmatique du terme créant la tension avec le foyer métaphorique, puis, d'autre part, au § 3.2, Prox qui est une approche stochastique des graphes lexicaux à partir de laquelle SLAM calcule l'axe paradigmatique du foyer de la métaphore. À la section 4, nous analysons les résultats d'une évaluation de SLAM réalisée sur un ensemble de métaphores tirées d'APPROX (Duvignau et Gaume, 2001-2004), protocole expérimental que nous présentons au § 4.1 et sur lequel s'appuie la procédure d'évaluation décrite au § 4.2.

---

6. Même en ayant accès au niveau conceptuel, pour résoudre une métaphore comme *« déshabiller une pomme »,* il faudrait également arriver à résoudre, à ce niveau conceptuel, le système d'équations $C_1$:POMME::DESHABILLER:$C_4$ par $C_1$ = PELER et $C_4$ = POUPEE, ce qui n'est pas une tâche simple à automatiser.

## 2. Cadre théorique

### 2.1. La métaphore analogique ou approximation sémantique

Il est établi depuis longtemps que, au cours du développement normal du lexique, les apprentis locuteurs produisent des énoncés d'allure métaphorique du type : « *un mille-pattes on dirait un peigne* », « *ballon* » pour LUNE, qui manifestent leur aptitude à rapprocher des objets qui relèvent de catégories distinctes sur la base de propriétés physiques ou fonctionnelles qu'ils partagent (Winner, 1979 et 1995 ; Gelman *et al.*, 1998 ; Bassano, 2000). En ce qui concerne le développement du lexique verbal, champ beaucoup moins étudié, on retrouve chez les apprentis locuteurs cette compétence analogique, essentielle dans le développement cognitif et langagier, qui consiste à rapprocher des actions différentes. La mise au jour de la production de ce type d'énoncés, considérés le plus souvent, dans la littérature, comme des surextensions erronées ou des métaphores, montre que ces locuteurs en situation normale du manque d'un mot sont dotés d'une flexibilité sémantico-cognitive fondamentale qui leur permet, dès le plus jeune âge, de pallier ce manque et d'assurer la communication avec leurs interlocuteurs. En effet, si l'on examine, non plus les seules caractéristiques linguistiques des approximations sémantiques produites mais aussi la capacité cognitive qui conditionne leur génération, se manifeste alors, dans le cadre d'énoncés bâtis à partir d'items relationnels (verbes, adjectifs, noms de partie…), une compétence analogique qui reflète la flexibilité cognitive dont ces locuteurs sont dotés : « *déshabiller une pomme* » pour [l'action de PELER une POMME] grâce à l'identification d'une analogie du type PELER:POMME::DESHABILLER:POUPEE. Ces approximations sémantiques, manifestations linguistiques de cette compétence cognitive, peuvent alors se définir comme la mise sous tension d'un terme lexical, soit par rapport à son contexte linguistique, soit par rapport à son contexte pragmatique, donnant ainsi lieu à deux types d'approximations (Duvignau, 2002) :

– **les approximations sémantiques interdomaines** comme quand la séquence « *elle déshabille une pomme* » est produite pour décrire l'événement [l'action de PELER une POMME] qui mobilise la distance entre le domaine sémantique VETEMENT associé au lexème approximatif « *déshabiller* » et le domaine sémantique VEGETAL associé à l'événement décrit [l'action de PELER une POMME]. La relation entre le lexème approximatif « *déshabiller* » et un lexème conventionnel « *peler* » est alors une relation de cohyponymie[7] interdomaine dans laquelle les cohyponymes sont en relation de synonymie même s'ils appartiennent à des champs sémantiques différents : VETEMENT ≠ VEGETAL (Duvignau, 2003 ; Duvignau *et al.*, 2007). Cette relation de cohyponymie interdomaine (comme entre « *déshabiller* » et « *peler* ») est le pendant linguistique de ce que Gentner (Gentner *et al.*, 2001) appelle, au niveau conceptuel, la relation d'identité partielle entre des prédicats relationnels (comme entre DESHABILLER et PELER) issus de domaines conceptuels distincts ;

---

7. Deux termes sont cohyponymes s'ils partagent un même hyperonyme.

– **les approximations sémantiques intradomaines** comme quand la séquence « *elle déchire la feuille* » est produite pour décrire l'événement [l'action de FROISSER du PAPIER] qui mobilise la distance entre le concept DECHIRER, associé au lexème approximatif « *déchirer* », et le concept FROISSER, associé à l'événement décrit [l'action de FROISSER du PAPIER]. Ici, le lexème approximatif produit, « *déchirer* », est en relation de cohyponymie avec l'item attendu « *froisser* », tout en pouvant appartenir au même domaine sémantique. Il ne s'agit donc plus d'une relation de synonymie mais plutôt d'une relation de troponymie[8] intradomaine qui entraîne une tension pragmatique du fait de l'inadéquation du verbe troponymique pour désigner l'événement concerné.

Peu de travaux ont donné naissance à des théories et des modèles capables de prendre en charge l'analogie mis à part le *Structure-Mapping* de Gentner, (Gentner, 1983) et le CST d'Indurkhya, (Indurkhya, 1987), la plupart se cantonnant à la métaphore (Lakoff, 1992 ; Gibbs, 1994 ; Glucksberg et Keyzar, 1990). Pour notre part, nous nous intéressons ici à la modélisation et à l'automatisation de la résolution lexicale de l'approximation sémantique interdomaine, manifestation lexicale de l'analogie conceptuelle.

**2.2. Les modèles computationnels de la métaphore**

Les modèles computationnels existant autour de la métaphore se placent pour la plupart au niveau conceptuel : le *Structure-Mapping Engine* de Falkenhainer (Falkenhainer *et al.*, 1989), l'AST d'Indurkhya (Indurkhya, 1987), le modèle KARMA de Narayanan (Narayanan, 1997). Quelques-uns, cependant, se situent comme le nôtre au niveau lexico-sémantique et concernent le traitement automatique des langues. Parmi ceux-ci, on distingue, d'une part, les modèles nécessitant des connaissances lexicales préalables annotées à la main, parfois complexes, et d'autre part, les modèles qui, à partir de corpus bruts ou de dictionnaires, font émerger des connaissances lexicales utiles au modèle sans recours à l'annotation humaine.

Parmi les modèles du premier type, on retrouve le modèle Met* de Dan Fass (Fass, 1991). Celui-ci se démarque des autres modèles par son enjeu : repérer dans un énoncé les métonymies, les métaphores, les erreurs et proposer une solution littérale pour les métaphores. Il s'appuie sur la théorie des *Preference Semantics* développée par (Wilks, 1978). Selon cette théorie, des contraintes syntaxiques et lexicales portent sur l'ensemble des lexèmes. C'est la violation de ces contraintes qui permet d'identifier une métaphore analogique et d'en proposer une solution. Met* nécessite pour cela la connaissance détaillée des relations entre lexèmes dans le système de la langue et des contraintes qui s'y appliquent. Il peut alors traiter tous les types de métaphores, y compris les métaphores à pivot verbal. Ainsi, dans l'exemple 1, après avoir repéré l'item lexical métaphorique – « *drink* » (« *boire* ») – Met* propose « *use* » (« *consommer* ») comme solution littérale de substitution.

---

8. La relation de troponymie entre deux verbes introduite par C. Fellbaum est une relation de manière (Fellbaum, 1999), par exemple entre « *marcher* » et « *courir* ».

**Exemple 1 :** « *My car drinks gasoline.* » (*Ma voiture boit de l'essence.*)

Comme le remarquent Perlerin *et alii* (Perlerin *et al.*, 2002), cette approche fait appel à *de grandes bases de connaissances sémantiques, fortement structurées, pour sa mise en œuvre*. Pour cette raison, ce type de modèles est difficilement exploitable en TAL.

D'autres modèles de ce type sont dédiés à des tâches utilisateurs spécifiques. Ceux-ci ne nécessitent pas une description lexico-sémantique préalable détaillée mais une description restreinte à des domaines sémantiques utiles à la tâche. Ainsi, Perlerin *et alii* (Perlerin *et al.*, 2002) utilisent un modèle de représentation lexicale, Anadia, dérivé de la sémantique interprétative de (Rastier, 1987). Ce modèle permet de décrire à l'aide d'un dispositif (base de données relationnelle) l'ensemble des lexies d'un domaine sémantique. Cette description repose sur l'attribution de traits sémantiques (ou sémèmes) qui permettent de distinguer, différentiellement, les lexies du domaine. Une projection de ce dispositif sur un corpus permet avec l'aide d'une visualisation macroscopique (graphiques) et microscopique (coloration des textes et informations détaillés) de détecter les isotopies et de visualiser les termes susceptibles d'être métaphoriques. L'introduction de sèmes partagés, dans une version améliorée d'Anadia, offre une visualisation des isotopies interdomaines. Les métaphores conceptuelles sont ainsi appréhendées et peuvent être interprétées grâce aux valeurs des sèmes partagés.

Ils saisissent les réalisations lexicales des métaphores conceptuelles, qu'elles soient prédicatives nominales ou à pivot verbal. Par exemple, pour une métaphore analogique entre le domaine de la Bourse et le domaine de la météorologie, ils saisissent dans l'exemple 2 les lexèmes métaphoriques « *accalmie* » et « *réchauffé* ».

**Exemple 2 :** « *L'accalmie intervenue ensuite sur le front monétaire [...] a cependant réchauffé l'ambiance à la corbeille.* »

Notons, que l'enrichissement du dispositif Anadia est réalisée de façon cyclique par l'utilisateur (amélioration du dispositif, projection des résultats, visualisation, amélioration…). Les auteurs sont conscients que cette démarche cyclique place leur modèle au plus près de la représentation des domaines sémantiques par l'utilisateur pour une tâche donnée mais qu'elle constitue également une limite dans l'utilisation du dispositif : il y a autant d'instanciations du modèle que d'utilisateurs.

Dans le cadre d'une tâche plus spécifique encore autour de la métaphore – identifier et comprendre les modes de conceptualisation métaphorique dans le domaine de la biomédecine – Vandaele *et alii* (Vandaele *et al.*, 2006) utilisent également des outils de traitement automatique des langues. Vandaele reprend la théorie de Lakoff et Johnson (Lakoff et Johnson, 1980) selon laquelle les métaphores se réalisent au niveau conceptuel. Elle cherche alors à étudier ces conceptualisations métaphoriques dans les écrits scientifiques de biomédecine, par l'analyse de leurs réalisations lexicales. Remarquant, à la suite des travaux de (Duvignau, 2003) la spécificité et l'importance des unités lexicales prédicatives telles que le verbe, dans la réalisation lexicale des conceptualisations métaphoriques, ce sont sur elles que se sont portées les analyses linguistiques automatisées. Partant du principe qu'une métaphore se réalise par la projection des actants du domaine source sur ceux du domaine cible, Vandaele *et alii* ont annoté manuellement des corpus en repérant (1) l'indice de conceptualisation métaphorique (2) la réalisation des actants dans la phrase et (3) les collocatifs. Une fois le corpus annoté, une extraction automatique permet d'analyser quantitativement puis qualitativement les modes de conceptualisation métaphorique. Vandaele *et alii* ont ainsi *cerné certains éléments clés de la conceptualisation métaphorique en science*. Cette dernière approche est intéressante de par son objectif : mieux comprendre, par une analyse lexicale et l'utilisation d'outils TAL, les processus cognitifs de niveau conceptuel, ici la conceptualisation métaphorique.

Notre modèle, quant à lui, est un modèle applicable en langue[9] et qui contrairement à celui de D. Fass ne nécessite aucune ressource annotée à la main. Nous ne connaissons à l'heure actuelle que deux modèles traitant de la métaphore analogique qui partage la deuxième caractéristique : le modèle de W. Kintsch (Kintsch, 2000) et le *Latent Relation Mapping Engine* (LRME) de P. Turney (Turney, 2008b). Le modèle de Kintsch partage également la première caractéristique contrairement à celui de Turney qui, comme nous allons le voir, nécessite une définition des termes lexicaux[10] pour chaque domaine sémantique mis en jeu, ce qui en limite l'application.

Dans son modèle, Kintsch propose une méthode automatique pour situer un énoncé, de type métaphorique ou littéral, dans un espace sémantique multidimensionnel. Pour cela, il utilise l'analyse sémantique latente – ou LSA pour *Latent Semantic Analysis* (pour une introduction à la LSA *cf.* T. K. Landauer *et al.*, 1998) qui permet de générer un espace sémantique multidimensionnel à $n$ dimensions (ici 300) dans lequel sont placés des segments linguistiques (mots, phrases, paragraphes…) extraits du corpus. Kintsch s'appuie sur la métrique de cette espace vectoriel pour récupérer l'ensemble des $m$ termes les plus proches du prédicat, terme source de la métaphore ($500 < m < 1\,500$). Il ne conserve alors que ceux qui sont assez fortement corrélés avec l'argument[11]. Le sens de l'énoncé de type métaphorique est alors représenté dans l'espace sémantique par le barycentre entre ces termes et le terme approximatif. Notons cependant que ce modèle a été évalué par Kintsch sans procédure rigoureuse et sur un nombre limité d'énoncés (7) contenant chacun une métaphore prédicative nominale du type de l'exemple 3. Cette lacune a en partie été comblée par (Bentsen et Cabiaux, 2002) qui ont adapté l'algorithme de Kintsch sur le français et effectué une évaluation rigoureuse sur 20 énoncés métaphoriques extraits de 9 contes de Maupassant. Ces énoncés comportent des métaphores variées dont des métaphores à pivot verbal comme dans l'exemple 4. Les résultats obtenus semblent valider la pertinence du modèle de Kintsch.

---

9. SLAM peut s'appliquer à un corpus général mais aussi à un corpus ciblé sur un domaine (voir § 5.2 Variabilité selon le choix du corpus K).
10. Un des objectifs de Turney est d'arriver à se passer de ces définitions.
11. Pour cela, il calcule automatiquement un seuil minimal de corrélation.

**Exemple 3 :** « *My lawyer is a shark* » (*Mon avocat est un requin*)

**Exemple 4 :** « *La lune verse une pluie de lumière* »

De son côté, le LRME de Turney repose sur la notion d'analogie lexicale. L'analogie lexicale $m_1:m_2::m_3:m_4$ entre les lexèmes $m_1$, $m_2$, $m_3$, $m_4$ est telle que, sémantiquement $m_1$ est à $m_2$ ce que $m_3$ est à $m_4$. Tous couple lexical $a_i:a_j$ où $a_i$ et $a_j$ appartiennent à un domaine sémantique A peut ainsi être mis en relation avec un couple $b_k:b_t$ où $b_k$ et $b_t$ appartiennent à un domaine sémantique B.

(Turney, 2006) a conçu la *Latent Relational Analysis* (LRA) qui est une méthode permettant de mesurer la similarité entre deux couples lexicaux (ou similarité relationnelle) : plus cette mesure est élevée, plus la relation lexicale entre les couples mis en jeu est de type analogique.

Le LRME utilise une version simplifiée de la LRA pour établir la meilleure bijection possible entre n lexèmes d'un domaine sémantique A et n lexèmes d'un domaine sémantique B. Le principe est le suivant : pour toute bijection $f$ entre le domaine sémantique A et le domaine sémantique B, plus la somme des mesures de similarité entre l'ensemble des couples $a_i:a_j$, $a_i$ et $a_j$ appartenant à A, et leurs images $f(a_i):f(b_j)$ dans B est élevée, meilleure est la bijection $f$.

Pour calculer la similarité entre les couples lexicaux de A et ceux de B, la première étape du LRME est de lister l'ensemble des couples $a_i:a_j$ issus de A et $b_i:b_j$ issus de B. Il extrait ensuite l'ensemble des séquences linguistiques de la forme *[0 à 1 mot] $a_i$ [0 à 3 mots] $a_j$ [0 à 1 mot]* (idem pour $b_i$ et $b_j$) dans un corpus du Web anglais[12]. Durant la troisième étape, il génère une liste de patrons à l'intérieur desquels $a_i$ et $a_j$ sont remplacés par *X* et *Y*, les mots restants étant soit substitués par un astérisque (signifiant que n'importe quel terme peut le remplacer) soit laissés tels quels. Ainsi la séquence « *le maçon coupe la pierre avec* » dans laquelle « *maçon* » et « *pierre* » sont les termes en relation va générer les patrons « *le X coupe * Y avec* », « * X * la Y * » »… (Turney, 2008a). Le nombre d'occurrences de chaque couple lexical au sein de chacun des patrons récupérés est stocké dans une matrice M (une ligne par couple lexical, une colonne par patron). La normalisation des effectifs et la réduction après une analyse factorielle[13] de la matrice M engendre la matrice R (une ligne par couple lexical). La mesure de similarité relationnelle entre deux couples correspond alors au cosinus entre les lignes correspondantes dans R et préalablement normalisées.

La pertinence des bijections calculées par ce modèle a été évaluée par (Turney, 2008b) à partir 10 métaphores conceptuelles fournies par (Lakoff et Johnson, 1980) et

---

12. Corpus extrait par Charles Clarke, Univestity of Waterloo.
13. Le LRME utilise la décomposition en valeurs singulières.

de 10 analogies scientifiques conceptuelles provenant de (Holyoak et Thagard, 1989). À l'issue de cette évaluation, la précision du LRME a été de 91,5 %.

Notons enfin que si notre modèle SLAM se rapproche sur certains points de ceux de Kintsch et Turney (efficacité en langue avec le premier, pas de ressource annotée manuellement avec les deux), il diverge sur deux points :

– la visée du modèle : la recherche d'une solution, avec la substitution de l'item lexical métaphorique par un item conventionnel ;

– la prise en compte de la syntaxe pour le traitement linguistique de la métaphore[14].

Ces deux points seront explicités dans la partie suivante qui décrit le modèle SLAM.

## 3. SLAM : solution lexicale automatique pour métaphore

Nous allons ci-dessous commencer par décrire brièvement Syntex puis Prox ce qui nous permettra ensuite de définir SLAM.

### 3.1. Syntex : pour extraire les triplets syntaxiques d'un corpus

Le corpus que nous avons utilisé pour extraire les triplets syntaxiques est le corpus Frantext.20 qui est issu de la base Frantext[15], de l'ATILF[16]. Il est composé de 515 romans du XX$^e$ siècle et comporte environ 30 millions de mots.

Frantext.20 a d'abord été étiqueté morphosyntaxiquement par l'outil Treetagger, développé à l'Université de Stuttgart, puis analysé syntaxiquement par l'outil Syntex[17], développé au sein de l'ERSS par Didier Bourigault (Bourigault *et al.*, 2005). Syntex est un analyseur syntaxique de corpus, qui prend en entrée un corpus de phrases étiquetées, et calcule pour chaque phrase les relations de dépendance syntaxique entre les mots (sujet, complément d'objet, complément prépositionnel, épithète, etc.) À partir de l'analyse syntaxique, sont extraits des triplets <régisseur, relation, régi>. Par exemple, de l'analyse syntaxique de la phrase « *il mange la souris* » est extrait le triplet *<V.manger, obj, N.souris>*. Au cours de cette étape d'extraction de triplets, un certain nombre de normalisations syntaxiques sont effectuées :

– intégration de la préposition : « *Il mange avec les doigts* » → *<V.manger, avec, N.doigt>* ;

---

14. Ce point a été évoqué par (Turney, 2008b) mais n'a pas encore été mis en œuvre dans son modèle.
15. http://atilf.atilf.fr/frantext.htm
16. http://www.atilf.fr/
17. http://w3.erss.univ-tlse2.fr/textes/pagespersos/bourigault/syntex.html

– distribution de la coordination : « *Il mange la pomme et la poire* » →
*<V.manger, obj, N.pomme>*, *<V.manger, obj, N.poire>* ;

– traitement du passif : « *la pomme a été mangée* » → *<V.manger, obj, N.pomme>* ;

– traitement de l'antécédence relative : « *Jean qui dort* » → *<V.dormir, suj, NP.jean>*.

L'analyseur syntaxique Syntex, nous a donc permis de construire une base de données de l'ensemble des triplets syntaxiques *<régisseur, relation, régi>* présents dans Frantext.20 ainsi que leurs effectifs.

**3.2. Pour calculer des similarités sur les graphes lexicaux**

Dans la suite, un graphe $G = (V, E)$ est défini par un ensemble $V$ de $n$ sommets, et un ensemble de $m$ arcs $E \subset V^2$. Dans ce papier $V$ est un ensemble de mots et l'ensemble des arcs $E$ est défini par une relation[18] $R$, $V \mapsto V$ : $(m_1, m_2) \in E$ si et seulement si $m_1 \xrightarrow{R} m_2$.

La plupart des graphes lexicaux, comme la majorité des graphes de terrain[19], sont des *Small World* (SW) (Watts et Strogatz, 1998 ; Ravasz et Barabási, 2003 ; Gaume, 2004), et possèdent des propriétés structurelles bien particulières :

– **P1) la densité en arcs est faible :** les SW sont peu denses en arcs, en général, $m = O(n \log(n))$ ;

– **P2) la moyenne des plus courts chemins est petite :** dans les SW la moyenne des plus courts chemins entre sommets $(L)$ est petite, en général il existe au moins un chemin court entre n'importe quelles paires de sommets ;

– **P3) il existe des Clusters :** dans les SW, le taux de 'clustering' $(C)$ qui exprime la probabilité qu'ont deux sommets distincts d'être adjacents s'ils sont adjacents à un troisième sommet est d'un ordre de magnitude plus grand que pour les graphes aléatoires à la Erdos-Renyi : $C_{SW} \gg C_{Random}$ ; ce qui indique que les SW sont localement denses en arcs, alors que globalement leur densité est faible (propriété P1) ;

– **P4) la distribution des degrés est à queue lourde :** la distribution du degré d'incidence des sommets suit une loi de puissance. La probabilité $P(k)$ qu'un sommet donné ait $k$ voisins décroît comme une loi de puissance : $P(k) = k^{-a}$ ($a$ étant une constante caractéristique du graphe), alors que c'est une loi de Poisson dans les graphes aléatoires.

---

18. Dans les graphes sémantiques, $R$ peut être une relation *syntagmatique* (Ide et Véronis, 1998 ; Karov et Edeman, 1998 ; Lebart et Salem, 1994), ou une relation *paradigmatique* (Fellbaum, 1999 ; Ploux et Victorri, 1998).

19. Les graphes de terrain sont les graphes que l'on trouve en pratique. Ils sont construits à partir de données de terrain. On les retrouve dans toutes les sciences de terrain. Par exemple, le graphe des collaborations scientifiques (les sommets sont les auteurs d'articles scientifiques, et on relie deux auteurs A et B s'ils ont au moins une publication en commun).

Les propriétés $P_3$ (aspect communautaire) et $P_4$ (aspect hiérarchique) révèlent des phénomènes fondamentaux dont sont issues ces structures, permettant ainsi une meilleure compréhension et exploitation des données représentées par ces graphes. Il est remarquable que la plupart des graphes de terrain, dont les graphes lexicaux, se ressemblent tous par leur structure commune de SW, bien qu'intrinsèquement cette structure soit très rare du point de vue de la théorie de la mesure (c'est-à-dire que la probabilité de tirer au hasard parmi l'ensemble des graphes possibles un graphe possédant ces quatre propriétés est très proche de zéro). Aussi, c'est dans le but de proposer des outils novateurs, bien adaptés à la structure des SW car s'appuyant sur leurs propriétés fondamentales, qu'est développé Prox[20].

**3.3. Prox**

Nous présentons ci-dessous le principe de base de Prox qui est le cœur d'une approche stochastique à partir de laquelle on peut construire de nombreux outils pour l'étude de la structure des SW (métrologie : Gaume, 2004 ; Gaume et Mathieu, 2009 ; visualisation : Gaume, 2008 ; génération automatique : Gaume, Mathieu, 2008 ; modélisation en psycholinguistique : Gaume, Duvignau, Prevot, Desalle, 2008).

Bien que dans un SW, le nombre d'arcs parcourus par un plus court chemin entre deux sommets $r$ et $s$ quelconques soit en général petit, il existe deux configurations opposées :

– **configuration 1** : les sommets $r$ et $s$ peuvent être reliés par un grand nombre de chemins courts ($r$ et $s$ sont fortement reliés : il existe une confluence forte de chemins entre $r$ et $s$) ;

– **configuration 2** : les sommets $r$ et $s$ peuvent être reliés par seulement quelques chemins courts ($r$ et $s$ sont faiblement reliés : faible confluence entre $r$ et $s$).

Bien entendu toutes les configurations intermédiaires peuvent exister : les sommets $r$ et $s$ peuvent être reliés par un nombre moyen de chemins courts ($r$ et $s$ sont moyennement reliés : il existe une confluence moyenne de chemins entre $r$ et $s$).

Une bonne notion de proximité entre sommets devrait donc rapprocher les paires de sommets en configuration 1 (forte confluence entre $r$ et $s$) et éloigner celles en configuration 2 (faible confluence entre $r$ et $s$).

Supposons que nous disposions d'un graphe connexe, symétrique et réflexif, $G = (V, E)$ de $n = |V|$ sommets et $m = |E|$ arcs, et que sur ce graphe une particule puisse à chaque instant $t \in \mathbb{N}$ se balader[21] aléatoirement sur ses sommets :

---

20. http://Prox.irit.fr

21. Du point de vue mathématique le processus stochastique que nous décrivons ici est une chaîne de Markov. Si nous notons $A = (a_{ij})$ la matrice d'adjacence du graphe ($a_{ij} = 1$ si $\{i,j\} \in E$ : i et j sont reliés par une arête ; $a_{ij} = 0$ si $\{i,j\} \notin E$ : i et j ne sont pas sont reliés par

– à l'instant *t* la particule est sur un sommet $r \in V$ ;

– quand la particule est à un instant *t* sur un sommet $u \in V$, elle ne peut atteindre à l'instant $t+1$, que les sommets $s \in V$ tels que $\{u,s\} \in E$ (c'est-à-dire les voisins de *u*). La particule se déplace de sommet en sommet à chaque instant en empruntant les arcs du graphe. On supposera de plus que pour tout sommet $u \in V$, chacun des arcs sortants de *u* sont équiprobables.

Si à l'instant de départ $t = 0$, la particule est sur le sommet $r \in V$ alors après *k* transitions, (c'est-à-dire à l'instant $t = k$), tout sommet $s \in V$ situé à une distance de *k* arcs ou moins de *r* peut être atteint. La probabilité qu'en partant à $t = 0$ du sommet *r*, la particule soit au temps *t* sur le sommet *s* dépendant de *t* et du nombre de chemins entre le sommet de départ *r* et le sommet *s*, de leurs longueurs et de la structure du graphe autour des sommets intermédiaires le long des chemins (plus il y a de chemins, plus ces chemins sont courts, et plus le degré des sommets intermédiaires est faible, plus la probabilité d'atteindre *s* depuis le sommet de départ *r* au *t-ième* pas est grande quand *t* reste petit). Au début de sa balade à partir d'un sommet de départ, la particule passe plus probablement sur les sommets vers lesquels le sommet de départ entretient une forte confluence, rapprochant ainsi les sommets d'un même cluster. Nous avons donc là une bonne notion de proximité entre sommets.

**Définition 1 :** $P_{G,\lambda,r}$

Pour un graphe $G = (V,E)$ et un sommet donné $r \in V$, nous noterons la fonction $P_{G,\lambda,r} : V \to [0,1]$, où $P_{G,\lambda,r}(s)$ est la probabilité d'atteindre le sommet *s* au temps $t = \lambda$ quand la balade aléatoire commence sur le sommet *r* au temps $t = 0$.

Pour tous sommets *r*, *s*, *u*, si $P_{G,\lambda,r}(s) > P_{G,\lambda,r}(u)$ alors ceci nous indique que la confluence de chemins entre *r* et *s* est plus forte que la confluence de chemins entre *r* et *u*.

Il nous sera utile par la suite de ranger dans une suite ordonnée tous les sommets d'un graphe par rapport à leur confluence relative depuis un sommet donné *r*, afin d'en extraire les $\gamma$ premiers mieux classés (les $\gamma$ sommets entretenant les plus fortes confluences depuis le sommet de départ *r*). Pour cela nous devons donner deux définitions :

**Définition 2 :** suite ordonnée selon un ordre inversé induit par une fonction *f*.

Soit A un ensemble fini et une fonction $f : A \to \mathbb{R}$

Posons alors la suite ordonnée $u_1, u_2, ..., u_n$ définie par :

(1) $\{u_1, u_2, ..., u_n\} = A$

---

une arête), la matrice de transition $M = (m_{ij})$ associée à la balade aléatoire de la particule est définie par $m_{ij} = a_{ij}/d(i)$ où $d(i)$ est le degré du sommet *i*, c'est-à-dire le nombre de voisins de *i* : $d(i)$ est égale à la somme de la *i-ème* ligne de la matrice *A*.

(2) $\forall i, k \in [1,n]$ si $i \leq k$ alors $f(u_i) \geq f(u_k)$

Nous dirons que *la suite $u_1, ..., u_n$ est la suite sur A selon un ordre inversé induit par f.*

Dans un graphe $G = (V, E)$, pour un sommet donné $r \in V$, l'ordre inversé induit par la fonction $P_{G,\lambda,r}$ sur $V$, range donc tous les sommets du graphe $G$ en fonction de leurs similarités décroissantes avec le sommet $r$ calculées par Prox au temps $\lambda$. Nous parlerons alors de *rang proxémique de x par rapport à r* pour parler du rang de $x$ selon l'ordre inversé induit par $P_{G,\lambda,r}$ sur $V$.

**Définition 3 :** $DIAM_{G,\lambda}(r, \gamma)$

Soit $G = (V, E)$ un graphe lexical et deux entiers non nul $\lambda$ et $\gamma \in \mathbb{N}^*$

Soit un lexème $r \in V$

Posons alors $DIAM_{G,\lambda}(r, \gamma) = \{X_1, X_2, ..., X_\gamma\}$ où les $X_1, X_2, ..., X_\gamma$ sont les $\gamma$ premiers éléments de la suite sur $V$ selon l'ordre inversé induit par $P_{G,\lambda,r}$.

**Exemple 5:** la figure 2 illustre $DIAM_{DicoSyn.Verbe,3}$(« *écorcer* »,*100*) qui est donc la liste des 100 sommets qui entretiennent les plus fortes confluences avec le verbe « *écorcer* » (du mieux classé : forte confluence avec « *écorcer* » – aux moins bien classés : moins forte confluence avec « *écorcer* ») calculée par Prox à $t = 3$ sur DicoSyn.Verbe[22].

---

22. DicoSyn.Verbe est un graphe construit à partir d'un dictionnaire de synonymes constitué de sept dictionnaires classiques : Bailly, Benac, Du Chazaud, Guizot, Lafaye, Larousse et Robert, dont ont été extraites les relations synonymiques. DicoSyn.verbe, qui est un SW typique, a 9 043 sommets et 110 939 arcs, sa plus grande partie connexe possède 8 993 sommets, sur laquelle son $L = 4,19$, son $C = 0,41$, sa courbe d'incidence est en loi de puissance de *Coefficient a* = -2,03 avec une *Corrélation* = 0,96.

> 1 →*écorcer*, 2 →*dépouiller*, 3 →*peler*, 4 →*tondre*, 5 *ôter*, 6 *éplucher*, 7 *raser*, 8 *démunir*, 9 →*décortiquer*, 10 *égorger*, 11 *écorcher*, 12 *écaler*, 13 *voler*, 14 *tailler*, 15 *râper*, 16 *plumer*, 17 *gratter*, 18 *enlever*, 19 *désosser*, 20 *déposséder*, 21 *couper*, 22 *bretauder*, 23 →*inciser*, 24 →*gemmer*, 25 →*démascler*, 26 →*baguer*, 27 *évincer*, 28 *étriller*, 29 *étrangler*, 30 *épurer*, 31 *émonder*, 32 *écailler*, 33 *ébrancher*, 34 *ébourrer*, 35 *ébarber*, 36 *tamiser*, 37 *taillader*, 38 *spolier*, 39 *sevrer*, 40 *scruter*, 41 *scarifier*, 42 *saler*, 43 *saigner*, 44 *s'épolier*, 45 *révoquer*, 46 *ruiner*, 47 *retourner*, 48 *retirer*, 49 *rançonner*, 50 *raisonner*, 51 *quitter*, 52 *priver*, 53 *piller*, 54 *perdre*, 55 *ouvrir*, 56 *nettoyer*, 57 *monder*, 58 *marquer*, 59 *lire*, 60 *isoler*, 61 *gruger*, 62 *fusiller*, 63 *frustrer*, 64 *fouiller*, 65 *filouter*, 66 *faufiler*, 67 *faucher*, 68 *exproprier*, 69 *examiner*, 70 *estamper*, 71 *escroquer*, 72 *entamer*, 73 *entailler*, 74 *effeuiller*, 75 *dévêtir*, 76 *développer*, 77 *dévaster*, 78 *dévaliser*, 79 *détrôner*, 80 *détrousser*, 81 *déshériter*, 82 *déshabiller*, 83 *désenvelopper*, 84 *désencombrer*, 85 *désavantager*, 86 *dérober*, 87 *dépourvoir*, 88 *dépiauter*, 89 *dépecer*, 90 *dénuer*, 91 *dénuder*, 92 *dénantir*, 93 *démonétiser*, 94 *dégarnir*, 95 *dégager*, 96 *défeuiller*, 97 *défaire*, 98 *décérébrer*, 99 *découronner*, 100 *déchausser*, ...

**Figure 2.** $DIAM_{DicoSyn.Verbe,3}(écorcer, 100)$ **:** *les 100 verbes entretenant les plus fortes confluences depuis « écorcer » dans DicoSyn.Verbe*

Dans DicoSyn.Verbe, le sommet *« écorcer »* a 8 synonymes : {*baguer, décortiquer, démascler, dépouiller, gemmer, inciser, peler, tondre*}. Dans la figure 2, les voisins de *« écorcer »* sont précédés par une flèche→ et le nombre qui précède chaque verbe est son rang proxémique par rapport à *« écorcer »*: si un verbe $Y_1$ est classé *k-ième*, et un autre verbe $Y_2$ est classé *k +1-ième*, c'est parce que :

$$P_{DicoSyn.Verbe,3,écorcer}(Y_1) \geq P_{DicoSyn.Verbe,3,écorcer}(Y_2)$$

C'est-à-dire que quand la particule démarre sa balade à travers les arêtes du graphe DicoSyn.Verbe à l'instant *t = 0* sur le sommet *« écorcer »*, la probabilité que la particule soit à l'instant *t = 3* sur le sommet $Y_1$ est plus grande ou égale à la probabilité qu'elle soit sur le sommet $Y_2$ à l'instant *t = 3* (la confluence depuis *« écorcer »* vers $Y_1$ est plus grande ou égale à la confluence depuis *« écorcer »* vers $Y_2$).

Il est important de remarquer ici que bien que *«* →*décortiquer »* soit un voisin direct de *« écorcer »* et que *« éplucher »* ne le soit pas, *« éplucher »* (6$^e$) est cependant mieux classé que le verbe *« décortiquer »* (9$^e$). Cela est rendu possible grâce à notre classement des verbes selon leurs rangs proxémiques par rapport à *« écorcer »*, ce qui aurait été impossible en utilisant un classement selon la longueur topologique des plus courts chemins à *« écorcer »*. En effet avec un tel classement topologique on aurait d'abord trouvé les voisins de *« écorcer »* (qui sont à distance topologique 1 de *« écorcer »*), puis tous les sommets à distance 2, puis tous ceux à distance 3… On atteint rapidement tous les sommets puisque DicoSyn.Verbe est un SW avec une faible moyenne des plus courts chemins.

De plus, dans DicoSyn.Verbe, il arrive très souvent que deux sommets *u* et *v* soient à égale distance topologique de *« écorcer »* (donc indifférentiables relativement à *« écorcer »*) alors que le rang proxémique nous permet de différencier *u* et *v* grâce à leurs confluences distinctes relativement à *« écorcer »*. En effet, DicoSyn.Verbe est un SW avec une structure clusterisée où Prox rapproche les sommets d'un même cluster.

$P_{DicoSyn.Verbe,3,écorcer}$ la proxémie de *« écorcer »* calculée par l'algorithme Prox organise ainsi l'axe paradigmatique de *« écorcer »* dans un continuum (Duvignau et Gaume, 2005) et c'est cela qui va permettre à SLAM d'automatiser la construction d'une liste ordonnée $[S_1, S_2, ..., S_k]$ de *k* candidats solutions à une métaphore.

### 3.4. SLAM : pour résoudre les métaphores analogiques

Maintenant que nous avons tout en main pour croiser l'axe paradigmatique avec l'axe syntagmatique d'une métaphore, nous pouvons définir SLAM.

SLAM prend en entrée la métaphore analogique à résoudre, codée sous la forme d'un triplet syntaxique *<X, Y, Z>* où *X* est le régisseur (précédé de sa catégorie lexicale), *Y* est la relation et *Z* est le régi (précédé de sa catégorie lexicale) et fournit en sortie une liste ordonnée de *k* candidats solutions $[S_1, S_2, ..., S_k]$. Une étoile indique sur quel argument (régi ou régisseur) porte le foyer métaphorique qu'il faut résoudre :

– *<X\*, Y, Z>* si le foyer métaphorique à résoudre est le régisseur :

    *<V.casser\**, obj, *N.livre>*→SLAM→[*V.déchirer*, …]

Notons que les catégories lexicales des solutions fournies en sortie par SLAM sont contraintes à toujours être identiques à la catégorie lexicale de l'argument sur lequel porte le foyer métaphorique qu'il faut résoudre. D'autre part, si SLAM n'arrive pas à résoudre la métaphore analogique *<X\*, Y, Z>*, la liste vide [] est alors retournée en sortie :

    Soit *<X\*, Y, Z>* une métaphore analogique à résoudre
    Soit c(*X*) le concept[23] associé au terme X dans le contexte *<X, Y, Z>*
    Soit c(*Z*)[22] le concept associé au terme Z dans le contexte *<X, Y, Z>*
    Soit *S* un candidat solution : $S \in [S_1, S_2, ..., S_k]$ = *SLAM(< X\*,Y,Z >)*
    Soit c(*S*) le concept associé au terme *S* dans le contexte *<S, Y, Z>*.

Pour que le terme *S* soit une solution pertinente à la métaphore analogique *<X\*, Y, Z>* il faut que *S* satisfasse les deux conditions suivantes :

    **C₁ :** *S* est conventionnel dans le contexte du triplet *<S, Y, Z>*

    **C₂ :** ∃ un concept C tel que le quadruplet c(*S*):c(*Z*)::c(*X*):C est un quadruplet analogique.

---

23. On suppose ici que le contexte *<X, Y, Z>* suffit à désambiguïser le sens de *X* pour identifier le concept c(*X*) qui lui est associé, ce qui n'est pas toujours le cas (voir configuration 2 du § 4.3).

Par exemple : $X$ = *V.déshabiller* et son concept associé c($X$) = DESHABILLER, $Y$ = *obj*, $Z$ = *N.pomme* et son concept associé c($Z$) = POMME ;

Le terme $Z_1$ = *V.peler* (avec son concept associé PELER) serait une bonne solution pour résoudre <*V.déshabiller\*, obj, N.pomme*> car en posant C = POUPEE nous avons bien :

**C$_{1bis}$** : *V.peler* est conventionnel dans le triplet <*V.peler, obj, N.pomme*>

**C$_{2bis}$** : PELER:POMME::DESHABILLER:POUPEE est un quadruplet analogique.

Le terme $Z_2$ = *V.manger* (avec son concept associé MANGER) n'est pas une bonne solution pour résoudre <*V.déshabiller\*, obj, N.pomme*> car même si *V.manger* peut être conventionnel dans le triplet <*V.manger, obj, N.pomme*>, les quatre concepts MANGER, POMME, DESHABILLER, PERSONNE ne forment pas un quadruplet analogique, et il semble difficile d'imaginer un concept C tel que MANGER:POMME::DESHABILLER:C pour pouvoir satisfaire la condition $C_2$.

Pour résoudre la métaphore analogique <$X^*, Y, Z$>, il nous faut donc commencer par trouver un terme $S$ (de même catégorie lexicale que $X$) qui satisfasse la condition $C_1$. Or, si $S$ peut être conventionnelle dans le triplet <*S, Y, Z*>, on peut espérer que des occurrences de ce triplet <*S, Y, Z*> soient présentes dans un grand corpus comme Frantext.20 pour lequel nous disposons grâce à Syntex de l'ensemble des triplets de la forme <_, *Y, Z*> (voir § 3.1).

**Définition 4 :** $T(K, <X^*, Y, Z>, \alpha, \beta)$

Soit un corpus $K$, une métaphore analogique <$X^*, Y, Z$>, deux entiers naturels $\alpha$, $\beta \in \mathbb{N}^*$

Posons alors $T(K, <X^*, Y, Z>, \alpha, \beta)$ l'ensemble de termes définis par :

$u \in T(K, <X^*, Y, Z>, \alpha, \beta)$ ssi $u$ est de même catégorie lexicale que X et d'effectif $\leq \beta$ dans le corpus $K$ et que $\exists$ un triplet de la forme <*u, Y, Z*> d'effectif $\geq \alpha$.

Nous faisons alors l'hypothèse que s'il existe une solution conventionnelle $S$ à la métaphore <$X^*, Y, Z$> alors $S \in T(Frantext.20, <X^*, Y, Z>, 3, 15000)$. Nous posons le seuil d'effectif $\alpha$ égal à 3 afin de favoriser la conventionalité de $S$. En effet, si un triplet <*S, Y, Z*> apparaît au moins 3 fois dans Frantext.20 on peut alors espérer que $S$ soit conventionnel dans le contexte du triplet <*S, Y, Z*>. De plus, de manière à ce que $S$ ne soit pas un lexème « passe-partout », très fréquent en langue, tel que le verbe *V.faire*, nous définissons un seuil maximal sur l'effectif d'apparition de $S$ dans le corpus :15 000 occurrences pour Frantext.20.

Par exemple, dans le corpus Frantext.20, on trouve huit triplets de la forme <*V._, obj, N.orange*> dont l'effectif est supérieure à 3 :

7 fois <V.apporter, obj, N.orange> ; 6 fois <V.manger, obj, N.orange> ; 4 fois <V.saisir, obj, N.orange> ; 4 fois <V.acheter, obj, N.orange> ; 3 fois <V.éplucher,

obj, N.orange> ; 3 fois <V.sentir, obj, N.orange> ; 3 fois <V.donner, obj, N.orange> ; 3 fois <V.voir, obj, N.orange>.

Puisque dans la suite de cet article nous fixons le corpus $K = Frantext.20$ et les seuils $\alpha = 3$, $\beta = 15000$, afin d'alléger les notations nous noterons :

$$T(<X, Y, Z>) \text{ pour } T(Frantext.20, <X, Y, Z>, 3, 15000)$$

Si nous voulons résoudre la métaphore *<V.déshabiller\*, obj, N.orange>*, il nous faut donc maintenant sélectionner parmi les huit éléments[24] de *T(<V.déshabiller\*, obj, N.orange>) = { V.apporter, V.manger, V.saisir, V.acheter, V.éplucher, V.sentir, V.donner, V.voir }* les éléments qui seraient susceptibles d'être des bons candidats solutions, c'est-à-dire qui satisfont la condition $C_2$.

Nous pouvons d'ores et déjà remarquer dans cet exemple que *V.éplucher* qui appartient à *T(<V.déshabiller\*, obj, N.orange>)* est une solution appropriée puisque *V.éplucher* satisfait $C_2$. En effet, EPLUCHER:ORANGE::DESHABILLER:PERSONNE est un quadruplet analogique. Cependant le triplet syntaxique *<V.éplucher, obj, N.orange>* apparait 3 fois dans le corpus Frantext.20 et n'est pas le plus fréquent des triplets de la forme *<V.\_, obj, N.orange>*, le plus fréquent étant *<V.apporter, obj, N.orange>* qui apparaît 7 fois. Or, les quatre concepts APPORTER, ORANGE, DESHABILLER, PERSONNE ne forment pas un quadruplet analogique, et il semble difficile d'imaginer un concept C tel que APPORTER:ORANGE::DESHABILLER:C. Il paraît donc peu probable de pouvoir satisfaire la condition $C_2$. Cela peut s'expliquer par une distance conceptuelle trop grande entre APPORTER et DESHABILLER et donc par une distance proxémique trop grande entre *V.apporter* et *V.déshabiller*.

Choisir pour solution de la métaphore *<X\*, Y, Z>*, l'élément $E$ parmi les éléments de *T(<X\*, Y, Z>)* dont le triplet *<E, Y, Z>* a le plus grand effectif n'est donc pas une voie envisageable pour trouver un élément qui satisfasse la condition $C_2$ si l'on ne contraint pas la distance proxémique entre $X$ et $E$.

Partant de ce constat, il est nécessaire de filtrer *T(<X\*, Y, Z>)* de manière à ne conserver comme candidats à la solution que les lexèmes les plus proches du foyer métaphorique sur l'axe paradigmatique. Pour filtrer *T(<X\*, Y, Z>)*, nous pouvons utiliser $DIAM_{DicoSyn.Verbe,3}(X,\gamma)$ qui est l'ensemble des $\gamma$ lexèmes qui entretiennent les plus fortes confluences avec le foyer de la métaphore $X^*$ dans DicoSyn.Verbe, en construisant l'intersection : $T(<X^*,Y,Z>) \cap DIAM_{DicoSyn.Verbe,3}(X,\gamma)$. Ne sont donc conservés dans cette intersection que les candidats qui entretiennent les plus fortes confluences avec le foyer de la métaphore $X^*$.

De manière à favoriser la conventionalité de la solution $S$ pour une métaphore *<X\*, Y, Z>*, nous devons ordonner les candidats solution restants en liste *{ $S_1$, ..., $S_n$ }* selon l'ordre inversé induit par l'effectif des triplets *<$S_i$, Y, Z>* dans Frantext.20.

---

24. Ces huit verbes font donc partie de l'axe syntagmatique de *« orange »*.

Nous pouvons maintenant définir formellement les solutions lexicales automatiques pour la métaphore $<X^*, Y, Z>$ sous la forme d'une suite ordonnée : nous pourrons ainsi évaluer les résultats selon les méthodes à la TopRank (voir § 4.2).

**Définition 5 :** $SLAM(<X^*,Y,Z>,(K,\alpha,\beta),(G,\lambda,\gamma))$

Soit une métaphore analogique $<X^*, Y, Z>$

Soit un corpus $K$ et deux entiers non nuls $\alpha, \beta \in \mathbb{N}^*$

Soit $G = (V, E)$ un graphe lexical et deux entiers non nuls $\lambda, \gamma \in \mathbb{N}^*$

Posons alors $SLAM(<X^*,Y,Z>,(K,\alpha,\beta),(G,\lambda,\gamma)) = [X_1, X_2, ..., X_n]$ où

$\{X_1, X_2, ..., X_n\} = T(K, <X^*,Y,Z>, \alpha, \beta) \cap DIAM_{G,\lambda}(X, \gamma)$, en rangeant les éléments de la suite selon l'ordre inversé, induit par les effectifs des triplets $<X_i, Y, Z>$ dans Frantext.20.

Puisque dans la suite de cet article nous fixons le graphe $G = DicoSyn.Verbe$, la longueur des balades $\lambda = 3$, le corpus $K = Frantext.20$ et les seuils $\alpha = 3$, $\beta = 15000$, afin d'alléger les notations nous noterons : $SLAM(<X,Y,Z>, \gamma)$ pour : $SLAM(<X,Y,Z>,(Frantext.20, 3, 15000),(DicoSyn.Verbe, 3, \gamma))$

**Exemple 6:** $SLAM(<V.déshabiller^*, obj, N.orange>, 90) = V.éplucher$ où le rang proxémique de *V.éplucher* par rapport à *V.déshabiller* est le 85[e] et l'effectif du triplet $<V.éplucher, obj, N.orange>$ dans Frantext est de 3.

**Exemple 7:** $SLAM(<V.miauler^*, suj, N.porte>, 40) = V.grincer, V.craquer$ où le rang proxémique de *V.grincer* et *V.craquer* par rapport à *V.miauler* sont respectivement 10 et 28 et les effectifs des triplets $<V.grincer, suj, N.porte>$, $<N.craquer, suj, N.porte>$ dans Frantext.20 sont respectivement de 31 et 3.

**Exemple 8:** $SLAM(<N.bras^*, de, N.arbre>, 40)) = N.branche, N.force, N.puissance$ où le rang proxémique de *N.force, N.puissance, N.branche* par rapport à *N.bras* sont respectivement 27, 13, 21 et les effectifs des triplets $<N.branche, de, N.arbre>$, $<N.force, de, N.arbre>$, $<N.puissance, de, N.arbre>$, dans Frantext.20 sont respectivement de 117, 4 et 3.

Dans l'exemple 8 le filtre sur le rang proxémique paramétré à 40 permet de ne retenir que des lexèmes assez proches sémantiquement de *N.bras* pour entretenir un rapport de synonymie interdomaine avec *N.bras*. Le lexème *N.tronc* est ainsi rejeté malgré l'effectif élevé du triplet $<N.tronc, de, N.arbre>$ dans Frantext.20 : 206 occurrences. Le tri sur les effectifs des triplets restants $<N._, de, N.arbre>$, quant à lui, favorise la conventionalité de la réponse, préférant alors *N.branche* à *N.force* ou *N.puissance* qui apparaissent moins fréquemment que *N.branche* avec *N.arbre* en tant que régisseur selon la relation 'de' dans Frantext.20.

SLAM ayant été défini, la partie suivante présente une évaluation du modèle à partir de données extraites de la plate-forme Flexsem.

## 4. Évaluation

### 4.1. Données Flexsem et protocole Approx

Pour l'évaluation de SLAM, nous nous sommes appuyés sur un ensemble de données recueillies avec le support du ministère de la Recherche[25] sur la plate-forme Flexsem.

Les données de Flexsem sont issues du protocole Approx, réalisé dans le cadre d'un projet École & Cognitique (Duvignau, Gaume, 2001-2004) qui se compose d'un matériel vidéo et d'une tâche de dénomination d'action, éléments présentés ci-après :

– matériel : 17 séquences vidéo d'actions standardisées de 45 secondes qui renvoient à 3 catégories d'action (voir tableau 1) ;

– tâche : chacun des participants a été placé en situation de dénomination puis de reformulation d'action à l'oral à partir des 17 vidéos :

   - phase explicative : *« on va voir des petits films où une dame fait quelque chose. Quant elle aura fini je te/vous demanderai « qu'est-ce qu'elle a fait la dame ? ». Il faudra alors dire ce qu'elle a fait »*,

   - phase expérimentale : épreuve de dénomination et de reformulation (seule la tâche de dénomination est considérée ici),

   - tâche de dénomination donnée au moment où l'action est terminée avec son résultat visible : *« qu'est-ce qu'elle a fait la dame ? »*,

   - tâche de reformulation donnée en suivant : *« ce qu'elle a fait la dame, dis-le-moi d'une autre manière, avec d'autres mots. »*.

| [DETERIORER] | [ENLEVER] | [SEPARER] |
|---|---|---|
| Éclater un ballon<br>Froisser une feuille de papier<br>Casser un verre avec un marteau<br>Écraser une tomate avec la main<br>Déchirer un journal | Peler une carotte avec un éplucheur<br>Éplucher une orange avec les mains<br>Enlever l'écorce d'une bûche<br>Déshabiller un poupon<br>Démonter une structure en Lego<br>Éplucher une banane | Émietter du pain avec ses mains<br>Couper un pain avec un couteau<br>Couper un pain avec ses mains<br>Hacher du persil avec un couteau<br>Scier une planche en bois<br>Découdre une chemise |

**Tableau 1.** *Matériel expérimental du protocole Approx*

---

25. Projet ANR Gaume 2008 et Projet ACI Jeunes Chercheurs Duvignau 2004.

Les réponses au protocole saisies dans Flexsem sont qualifiées par des linguistes spécialistes d'une langue de passation[26] sur différents critères dont un qui juge de la conventionalité ou du type d'approximation des verbes produits dans les réponses valides pour l'action dénommée. Ce critère peut prendre les valeurs suivantes, les deux dernières correspondant aux définitions données au § 2 :

1) conventionnel ;

2) approximation intradomaine ;

3) approximation interdomaine.

À partir d'un extrait des données[27] de Flexsem pour le français, nous pouvons maintenant évaluer SLAM.

**4.2. Procédure d'évaluation**

À partir de notre corpus d'évaluation, extrait des données de Flexsem pour le français, nous avons pour chaque film d'action f :

– l'ensemble des verbes conventionnels : $V_c(f)$ ;

– l'ensemble des triplets métaphoriques : $T_m(f)$.

Soit n un entier, la tâche à réaliser par SLAM est de trouver, pour chacun des triplets métaphoriques $<v^*, obj, Z> \in T(f)$, un ensemble ordonné de n solutions, que nous noterons : $S_n(<v^*, obj, Z>)$, ensemble des solutions de rang $\leq n$ de $S(<v^*, obj, Z>, 40)$.

L'ensemble de solutions trouvées pour chaque triplet $\in T(f)$ est alors comparé à l'ensemble des verbes conventionnels pour le film d'action f, $V_c(f)$.

Ainsi, pour tout n, la précision et le rappel de SLAM sont égaux au nombre de triplets métaphoriques $<v^*, obj, Z> \in T(f)$ pour lesquels il existe au moins une solution SLAM conventionnelle par rapport :

– pour la précision : au nombre de triplets métaphoriques ayant au moins une solution SLAM ;

– pour le rappel : au nombre de triplets métaphoriques.

La f-mesure est la moyenne harmonique de la précision et du rappel.

---

26. Langue dans laquelle une passation d'Approx peut être effectuée.

27. 84 énoncés métaphoriques contenant une approximation sémantique par rapport à l'action à dénommer.

**4.3. Analyse des résultats**

Les résultats obtenus pour l'évaluation de SLAM à partir du corpus d'évaluation extrait de Flexsem sont les suivants :

|           | n = 1 | n = 2 | n = 3 |
|-----------|-------|-------|-------|
| Précision | 0,426 | 0,489 | 0,511 |
| Rappel    | 0,238 | 0,274 | 0,286 |
| f-mesure  | 0,305 | 0,351 | 0,366 |

**Tableau 2.** *Résultats de l'évaluation 1 de SLAM*

Voici deux exemples pour lesquels la solution trouvée par SLAM est conventionnelle, c'est-à-dire $\in V_c(f)$ :

**Exemple 9 :** $S_1(<V.déplumer*, obj, N.banane>) = V.éplucher$ pour le film [EPLUCHER BANANE]

**Exemple 10 :** $S_1(<V.casser*, obj, N.feuille>) = V.déchirer$ pour [DECHIRER JOURNAL]

*A priori*, ces résultats paraissent faibles. Nous ne connaissons pas d'évaluation du même type permettant une comparaison, mais d'autres évaluations existent comme, par exemple, celle utilisée par Turney (voir le modèle LRME au § 2.2). La précision moyenne obtenue par seize méthodes différentes de cette évaluation est de 61,4 %[28] (Turney, 2008b). Toutefois, la complexité de la tâche réalisée durant cette évaluation est plus faible que celle liée à notre évaluation. En effet, dans l'évaluation de Turney, les domaines sémantiques à faire correspondre sont prédéfinis avec, en moyenne, sept termes lexicaux par domaine. La conséquence est une nette réduction de l'ambiguïté sémantique :

(1) pas d'ambiguïté interdomaine, les domaines source et cible étant connus ;

(2) peu d'ambiguïté au sein d'un même domaine sémantique, le nombre de termes par domaine étant restreint.

Au contraire, l'évaluation de SLAM est effectuée en langue, sans aucune définition *a priori* des domaines sémantiques mis en jeu. Ces considérations faites, la précision de SLAM dans notre évaluation apparaît comme un premier résultat encourageant.

Nous verrons au § 4.3.1 qu'une redéfinition plus précise de notre corpus d'évaluation engendrera une augmentation significative de la précision et du rappel de SLAM.

---

28. L'écart type de 14,7 points. Seul le LRME se distingue des autres méthodes avec une précision de 91,4 %.

Nous distinguons, dans la suite de notre article, cinq configurations qui peuvent être à l'origine des échecs de SLAM (pas de solution ou solutions non conventionnelles au sens où elle n'appartiennent pas à $V_c(f)$).

4.3.1. *Configuration 1 : le foyer métaphorique du triplet métaphorique est en relation de troponymie avec les verbes conventionnels correspondants*

Comme nous l'avons définie au § 2.1, une approximation sémantique verbale d'allure métaphorique prend la forme d'un verbe en relation de synonymie interdomaine avec le verbe conventionnel attendu. Or, dans les approximations interdomaines telles que nous les avons définies, certains items lexicaux métaphoriques peuvent aussi entretenir un rapport de troponymie interdomaine avec le verbe conventionnel.

**Exemple 11 :** *« Elle a démoli la feuille »* pour [FROISSER FEUILLE] où $S_1$(<*V.démolir**, obj, *N.feuille*>) = *V.déchirer* (qui n'a pas été donné comme réponse conventionnelle pour le film [FROISSER FEUILLE]).

Dans cet exemple, *V.démolir* et *V.froisser,* verbe conventionnel pour [FROISSER FEUILLE], sont des co-hyponymes de *V.détériorer* associés respectivement aux concepts d'action FROISSER et DEMOLIR. Or, ces deux concepts, en plus d'appartenir à des domaines conceptuels différents, se distinguent par la manière et l'intensité avec lesquelles ils réalisent le concept DETERIORER. Ainsi, indépendamment de la manière dont sont réalisées les actions, DEMOLIR est à STRUCTURE ce que, FROISSER est à FEUILLE. Cet écart dans la manière engendre une tension pragmatique entre *V.démolir* et *V.froisser*. Il existe donc entre ces deux verbes une relation de troponymie (voir § 2.1 pour une définition de la troponymie) que l'on appellera troponymie interdomaine.

Résoudre une troponymie interdomaine, c'est résoudre, dans un premier temps, la tension sémantique due à l'appartenance à deux domaines sémantiques, puis, dans un second temps, la tension pragmatique due à la relation de troponymie. Or, SLAM est un modèle linguistique qui ne peut résoudre que la tension sémantique. C'est pourquoi, lorsque $S_1$(<*V.démolir**, obj, *N.feuille*>) a pour solution *V.déchirer*, seule la tension sémantique a été résolue : la tension pragmatique subsiste et *V.déchirer* est une approximation sémantique intradomaine de *V.froisser*.

Il est donc nécessaire de distinguer dans notre corpus d'évaluation les deux types d'approximations interdomaines mis au jour : les approximations interdomaines non troponymiques d'une part, les approximations interdomaines troponymiques d'autre part.

Nous avons ainsi repéré 20 approximations interdomaines troponymiques qui peuvent être réparties comme suit :

– sans solution par SLAM : 10 triplets :

  - <*V.déchirer*, obj, N.planche*> pour [SCIER PLANCHE] ;
  - <*V.briser*, obj, N.persil*> pour [HACHER PERSIL] ;
  - <*V.casser*, obj, N.orange*> pour [EPLUCHER ORANGE] ;

- <V.déboucher*, obj, N.ballon> pour [ECLATER BALLON] ;
- <V.déchirer*, obj, N.légo> pour [DEMONTER STUCTURE EN LEGO] ;
- <V.briser*, obj, N.écorce> pour [ECORCER ARBRE] ;
- <V.épuiser*, obj, N.orange> pour [EPLUCHER ORANGE] ;
- <V.pétrir*, obj, N.orange> pour [EPLUCHER ORANGE] ;
- <V.plier*, obj, N.tomate> pour [ECRASER TOMATE] ;
- <V.quitter*, obj, N.légo> pour [DEMONTER STRUCTURE EN LEGO] ;

– sans solution conventionnelle appartenant à $V_c(f)$ : 10 triplets :
- $S_1$(<V.briser*, obj, N.pain>) = V.couper pour [EMIETTER PAIN] ;
- $S_1$(<V.casser*, obj, N.livre>) = V.déchirer pour [FROISSER PAPIER] ;
- $S_1$(<V.casser*, obj, N.tomate>) = V.couper pour [ECRASER TOMATE] ;
- $S_1$(<V.casser*, obj, N.papier>) = V.déchirer pour [FROISSER PAPIER] ;
- $S_1$(<V.croiser*, obj, N.papier>) = V.serrer pour [FROISSER PAPIER] ;
- $S_1$(<V.déchirer*, obj, N.pain>)=V.couper pour [ECORCER ARBRE] ;
- $S_1$(<V.déchirer*, obj, N.tomate>) = V.couper pour [ECRASER TOMATE] ;
- $S_1$(<V.démolir*, obj, N.feuille>) = V.déchirer pour [FROISSER PAPIER] ;
- $S_1$(<V.émietter*, obj, N.arbre>) = V.couper pour [ECORCER ARBRE] ;
- $S_1$(<V.ouvrir*, obj, N.pain>) = V.couper pour [EMIETTER PAIN].

Rappelons que SLAM a résolu la tension sémantique de ces approximations et que les solutions proposées sont des approximations intradomaines : seule la tension pragmatique reste.

SLAM, modèle exclusivement linguistique, a pour objectif de résoudre la tension sémantique des triplets métaphoriques. Par conséquent, seuls les triplets métaphoriques contenant des approximations sémantiques interdomaines non troponymiques peuvent lui être fournis en entrée. Nous avons donc supprimé de notre corpus d'évaluation les 20 triplets listés ci-dessus. Les résultats obtenus sont maintenant les suivants :

|           | n = 1 | n = 2 | n = 3 |
|-----------|-------|-------|-------|
| Précision | 0,540 | 0,622 | 0,648 |
| Rappel    | 0,317 | 0,365 | 0,380 |
| f-mesure  | 0,400 | 0,460 | 0,480 |

**Tableau 3.** *Résultats de l'évaluation 2, après retrait des énoncés troponymiques*

La suppression du traitement par SLAM des 10 triplets sans solution entraîne une augmentation significative du rappel : + 7,9 % à n = 1 et + 9,4 à n = 3. De même, la suppression des 10 triplets sans solution conventionnelle entraîne une augmentation significative de la précision : + 11,4 % à n = 1 et + 13,7 % à n = 3.

4.3.2. Configuration 2 : l'objet du triplet métaphorique est polysémique

Une autre source d'erreur pour SLAM vient de la polysémie de l'objet $Z$ de certains triplets métaphoriques <v*,obj,Z> ∈ T(f). Ainsi, les triplets métaphoriques suivants sont non résolus : $S_1(<V.désarticuler*, obj, N.jeu>) = V.fausser \notin V_c(f)$ pour [DEMONTER STRUCTURE EN LEGO], $S_1(<V.éplucher*, obj, N.bois>) = V.fouiller \notin V_c(f)$ pour [ECORCER ARBRE].

Analysons le second cas pour lequel SLAM propose en première solution le verbe *V.fouiller*. Ici, *N.bois*, objet du triplet, peut faire référence à la matière (du bois) ou à l'ensemble d'arbres (le bois). Dans le film, l'action réalisée porte sur une bûche de bois. L'objet du triplet, *N.bois*, fait donc référence à la matière de la bûche. Or, *V.fouiller*, solution de SLAM, peut s'appliquer à un ensemble d'arbres (un bois) mais pas à du bois, matière de la bûche. La polysémie de *N.bois* (la matière ou l'ensemble d'arbres) entraîne une ambiguïté sémantique non gérée par SLAM qui ne tient compte ni du contexte pragmatique ni du co-texte hors du triplet. Remarquons, pour terminer, que *V.fouiller* pourrait être, dans un autre contexte, une solution conventionnelle pour le triplet *<V.éplucher*, obj, N.bois>*.

**Exemple 12 :** *Les policiers épluchent le bois à la recherche du moindre indice.*

4.3.3. Configuration 3 : l'objet du triplet métaphorique est une approximation lexicale de l'objet sur lequel porte l'action

Une troisième source d'erreur ou d'absence de réponse provient de la non-reconnaissance, par le participant, de l'objet sur lequel porte l'action. L'objet du triplet métaphorique produit est donc une approximation lexicale de l'objet sur lequel porte l'action. Nous avons repéré deux triplets métaphoriques non résolus qui entre dans cette configuration : $S_1(<V.casser*, obj, N.livre>) = V.déchirer$ pour [FROISSER PAPIER], $S_1(<V.déchirer*, obj, N.pain>) = V.couper$ pour [ECORCER ARBRE].

Prenons par exemple le triplet *<V.déchirer*, obj, N.pain>* pour [ECORCER ARBRE]. L'ensemble des actions que l'on effectue sur un arbre n'est pas le même que celui que l'on effectue sur du pain. L'approximation lexicale due à la non-reconnaissance de l'objet arbre (plus précisément une bûche de bois) se traduit dans SLAM par des solutions non conventionnelles (sauf hasard) car les résultats de Syntex sur l'axe syntagmatique pour *N.pain* ne correspondent pas à ceux de Syntex pour *N.arbre*.

Remarquons que le remplacement de ces triplets respectivement par *<V.casser*, obj, N.papier>* et par *<V.déchirer*, obj, N.arbre>* conduirait à la configuration 1 décrite ci-dessus.

4.3.4. Configuration 4 : l'ensemble des verbes conventionnels n'est pas assez grand

Il est possible que l'ensemble des verbes conventionnels produits par notre protocole expérimental Approx ne soit pas assez grand pour contenir le verbe solution de SLAM pourtant conventionnel.

**Exemple 13 :** *« Elle décolle l'écorce »* pour [ECORCER ARBRE] où $S_1(<V.décoller*, obj, N.écorce>) = V.arracher$.

Ici *V.arracher* est conventionnel pour l'action [ECORCER ARBRE]. Or, *V.arracher* n'a pas été produit pour le film [ECORCER ARBRE] : il ne fait donc pas partie de l'ensemble

V$_c$([ECORCER ARBRE]). Dans notre évaluation, *V.arracher* est donc compté à tort comme une solution erronée.

4.3.5. Configuration 5 : le corpus et/ou le graphe sont incomplets

L'incomplétude du corpus Frantext.20 sur lequel est appliqué SLAM engendre l'absence de réponse pour les verbes métaphoriques appliqués à un objet absent du corpus, comme *N.légo* dans $S_I(<V.déchirer^*, obj, N.légo>) = \varnothing$. De même, l'absence du verbe métaphorique dans DicoSyn.Verbe engendre une absence de réponse comme pour $S_I(<V.déniauquer^*, obj, N.légo>) = \varnothing$. Notre équipe est actuellement en train de travailler sur l'extraction d'un corpus du Web français, ce qui devrait limiter le problème lié au corpus. Prenons par exemple *<V.déchirer\*, obj, N.légo>*, *N.légo* n'apparaît ni dans le corpus Frantext ni dans le corpus des 10 ans du journal *Le Monde*. Pourtant, il est présent dans 47 400 000 pages sur le Web (Requête effectuée le 10 mars 2009). Ceci peut être expliqué par le fait que les corpus Frantext.20 et les 10 ans du *Monde* sont constitués chacun d'un type de textes spécifique alors qu'au contraire le corpus du Web contient un très grand nombre de types de textes. Une objection à notre optimisme peut cependant être formulée de la sorte : le Web est un corpus non organisé et certains types de textes peuvent être surreprésentés engendrant alors une surreprésentation de lexèmes spécifiques. Objection à laquelle nous répondrons en deux temps : notre première réponse est que le croisement dans SLAM de l'axe syntagmatique provenant du corpus avec l'axe paradigmatique devrait permettre un premier filtrage de ces lexèmes s'ils sont trop éloignés sémantiquement du verbe métaphorique ; la seconde, liée aux techniques d'aspiration du Web que nous utilisons, est explicitée au § 5.2.

## 5. Conclusion

SLAM prend en entrée une métaphore analogique *<X\*, Y, Z>* quelconque comme « *déshabiller\* une pomme* », et propose en sortie une solution lexicale conventionnelle *S* comme « *peler* », de manière automatique, par un croisement de l'axe paradigmatique du foyer de la métaphore (ex. : « *déshabiller\** ») avec l'axe syntagmatique du terme créant la tension avec le foyer métaphorique (ex. : « *pomme* »). Concernant l'axe paradigmatique du foyer de la métaphore *X\**, SLAM s'appuie sur Prox afin de définir une similarité entre entités lexicales sur un graphe de synonymes G pour construire $DIAM_{G,\lambda}(X,\gamma)$ : l'axe paradigmatique de *X* sur un rayon $\gamma$, et que nous noterons ici : $prag(G, X)$. Concernant l'axe syntagmatique du terme *Z* créant la tension avec le foyer métaphorique, SLAM s'appuie sur l'analyseur syntaxique Syntex afin d'extraire les triplets syntaxiques de la forme *<\_, Y, Z>* présents dans un corpus *K* avec leurs effectifs pour construire $T(K,<X^*,Y,Z>,\alpha,\beta)$ : la projection sur la relation *Y* de l'axe syntagmatique de *Z* calculé dans le corpus *K*, et que nous noterons ici : $synt(K,Z,Y)$. Notre hypothèse est que les solutions lexicales à la métaphore *<X\*, Y, Z>* appartiennent à $prag(G,X) \cap synt(K,Z,Y)$. SLAM range alors ces solutions potentielles selon l'ordre inversé des effectifs des triplets $<S_i, Y, Z>$ dans le corpus $K$ pour obtenir une suite ordonnée de termes : $S_1, S_2, ..., S_n$

### 5.1. Un pas vers la modélisation de l'interprétation de la métaphore vive

Nous pensons que le modèle présenté dans cet article est pertinent pour le TAL mais aussi pour la modélisation en psychologie cognitive. En effet SLAM peut être considéré comme un pas vers une modélisation des processus cognitifs mis en œuvre dans l'interprétation de la métaphore vive :

(1) détection de la tension sémantique créée par l'expression $<X, Y, Z>$ = « *déshabiller une pomme* » car une incongruité est ressentie[29] par la relation $Y = obj$ établie dans cette expression entre le concept $c(X)$ = DESHABILLER et le concept $c(Z)$ = POMME ;

(2) résolution de cette tension sémantique en exhibant deux concepts $c_1$ et $c_2$ tels que :

(2.1) $c_1$ est proche de $c(X)$ = DESHABILLER, et $c_1$ est associé à $c(Z)$ = POMME par la relation $Y = obj$ : exemple $c_1$ = PELER,

(2.2) $c(X)$ = DESHABILLER est associé à $c_2$ par la relation $Y = obj$ : exemple $c_2$ = POUPEE,

$c_1:c(Z)::c(X):c_2$ (PELER:POMME::DESHABILLER:POUPEE) formant ainsi un quadruplet analogique.

SLAM ne modélise au niveau linguistique que la phase (2.1) de ce processus cognitif. Aussi, nous poursuivons notre approche pour modéliser la phase (1) afin d'identifier au niveau linguistique les métaphores vives, et la phase (2.2) afin de les résoudre entièrement.

### 5.2. Variabilité selon le choix du corpus K

Dans cet article, nous nous sommes limités au cas où $G = DicoSyn.Verbe$ et $K = Frantext.20$, alors que nous avons vu au § 4.3.5 que SLAM est sensible au corpus $K$ utilisé, d'une part parce que certains mots peuvent être absents du corpus, et d'autre part parce que $synt(K,Z,Y)$ est fortement sensible au domaine du corpus.

Par exemple, si nous définissons $corp(m_1, m_2)$ le corpus construit de la manière suivante :

1) initialiser $corp(m_1, m_2)$ en récupérant sur Google un germe des 1 000 premières pages contenant les mots $m_1$ et $m_2$ ;

2) aspirer le Web à partir de ce germe en ne conservant dans $corp(m_1, m_2)$ que les pages contenant les mots $m_1$ et $m_2$ ;

---

29 Cette incongruité ressentie pouvant se détecter en mesurant le potentiel évoqué de type N400.

si $m_2 \neq m_3$, alors $synt(corp(m_1,m_2),Z,Y) \neq synt(corp(m_1,m_3),Z,Y)$.

Par exemple, pour résoudre la métaphore <V.déshabiller*, obj, N.arbre>, si SLAM utilisait le corpus *corp(« arbre », « jardin »)* alors on pourrait espérer que la réponse soit du type *V.tailler, V.déraciner, V.écorcer, V.déshabiller,* alors qu'avec le corpus *corp(« arbre », « écrou »)* la réponse serait *V.déposer, V.déshabiller*, (on trouve *V.déshabiller* dans les réponses de SLAM pour ces deux corpus car sur le Web on trouve des phrases du type « ... l'hiver déshabille l'arbre ... », « ... déshabiller l'arbre primaire de boîte de tous ses pignons... » dont certaines se retrouvent capturées dans nos corpus du type *corp(« arbre », _ ))*. Nous comptons développer cette piste car elle montre l'importance du choix des corpus en TAL (Habert *et al.*, 1997) qui, par exemple, peut être mis à profit par des approches de type SLAM pour la construction de terminologies sur des corpus spécialisés. Cette piste est aussi très intéressante pour la désambiguïsation du sens des mots qui doit dépendre non seulement du contexte syntaxique mais aussi du domaine d'application[30].

**5.3. Les limites de SLAM face aux données issues du protocole APPROX**

Dans cet article, nous avons évalué les performances de SLAM sur un ensemble de métaphores expérimentales tirées du protocole APPROX. Les échecs apparents de SLAM sont essentiellement dus au matériel expérimental qui nous a servi pour l'évaluation présentée dans cet article car, d'une part, il y manque des solutions conventionnelles (ce qui pénalise SLAM alors que le modèle donne une réponse pertinente) et, d'autre part, la couverture du corpus Frantexte.20 n'est pas adéquate face à la langue parlée utilisée dans le protocole APPROX. Aussi, nous comptons mener une campagne d'évaluation de SLAM[28] auprès d'un ensemble d'experts. Par ailleurs, l'analyse linguistique des configurations conduisant à un échec de la résolution de triplets métaphoriques par SLAM (absence de réponse ou réponse erronée) permet de saisir les qualités et limites du modèle, en particulier l'absence de résolution de la tension pragmatique. Elle nous amène également à réfléchir sur la définition des approximations dans la théorie linguistique et notamment à la définition de l'approximation interdomaine qui pourrait être déclinée en deux types : approximation interdomaine non troponymique et approximation interdomaine troponymique. D'autre part, l'étude des approximations sémantiques que nous développons est dotée d'une dimension translinguistique. Dans ce cadre nous travaillons actuellement à une comparaison du français et du mandarin (projet M3 : ANR Gaume 2008, Franco-Taiwanais). Cet article se situe donc au cœur du TAL. En effet, la théorie linguistique permet de réaliser et d'améliorer un outil computationnel de modélisation de la langue et, en contrepartie, cet outil permet un retour sur la théorie en vue de la faire avancer et de lui permettre de décrire plus finement les phénomènes considérés.

---

30 voir http://erss.irit.fr/slam où SLAM est accessible en ligne sur différents corpus.

## 6. Bibliographie